\documentclass[11pt,a4paper]{article}
\usepackage[hyperref]{naaclhlt2019}
\usepackage{times}
\usepackage{latexsym}
\usepackage{amsmath}
\usepackage{amssymb}
\usepackage{xspace}
\usepackage{makecell}
\usepackage{multirow}
\usepackage{subcaption}
\usepackage{url}
\usepackage{multicol}
\usepackage{multirow}
\usepackage{enumitem}
\usepackage{booktabs}
\usepackage{graphicx}  
\usepackage{algorithm,algorithmicx,algpseudocode}
\DeclareMathOperator*{\argmin}{arg\!min\xspace}

\newcommand{\Lagr}{\mathcal{L}}

\newcommand{\onehot}{\textit{OneHot}}
\newcommand{\expect}[1]{\ensuremath{\underset{#1}{\mathbb{E}}\xspace}}
\newcommand{\nop}[1]{}

\aclfinalcopy 
\def\aclpaperid{238} 

  {\list{}{\leftmargin=#1\rightmargin=#1}\item[]}%
  {\endlist}
  
\title{How Large a Vocabulary Does Text Classification Need? \\A Variational Approach to Vocabulary Selection}

\author{Wenhu Chen$^{\dagger}$, Yu Su$^\P$, Yilin Shen$^{\ddag}$, Zhiyu Chen$^{\dagger}$, Xifeng Yan$^{\dagger}$, William Wang$^{\dagger}$\\
University of California, Santa Barbara$^{\dagger}$\\
Ohio State University, Columbus$^\P$\\
Samsung Research, Moutain View$^{\ddag}$\\
{\tt\small \{wenhuchen,zhiyuchen,xyan,william\}@cs.ucsb.edu su.809@osu.edu yilin.shen@samsung.com}
}
\date{}

\begin{document}
\maketitle
\begin{abstract}

With the rapid development in deep learning, deep neural networks have been widely adopted in many real-life natural language applications. Under deep neural networks, a pre-defined vocabulary is required to vectorize text inputs. The canonical approach to select pre-defined vocabulary is based on the word frequency, where a threshold is selected to cut off the long tail distribution. However, we observed that such a simple approach could easily lead to under-sized vocabulary or over-sized vocabulary issues. Therefore, we are interested in understanding how the end-task classification accuracy is related to the vocabulary size and what is the minimum required vocabulary size to achieve a specific performance. In this paper, we provide a more sophisticated variational vocabulary dropout (VVD) based on variational dropout to perform vocabulary selection, which can intelligently select the subset of the vocabulary to achieve the required performance. To evaluate different algorithms on the newly proposed vocabulary selection problem, we propose two new metrics: Area Under Accuracy-Vocab Curve and Vocab Size under X\% Accuracy Drop. Through extensive experiments on various NLP classification tasks, our variational framework is shown to significantly outperform the frequency-based and other selection baselines on these metrics.
\end{abstract}

\begin{table}[thb]
\small
\centering
\begin{tabular}{llllll} 
\toprule
\begin{tabular}[c]{@{}l@{}} Cutoff\\Freq. \end{tabular} & Vocab & \begin{tabular}[c]{@{}l@{}}Remain\\Vocab \end{tabular} & \#Emb & \#CNN & \begin{tabular}[c]{@{}l@{}}\#Emb \\Ratio\end{tabular}  \\ 
\midrule
1                                                       & 60K  & 100\%                                                  & 15M   & 0.36M & 97.6\%                                                 \\ 
\midrule
5                                                       & 40K   & 21.7\%                                                 & 10M   & 0.36M & 95.6\%                                                 \\ 
\midrule
10                                                      & 24K   & 13\%                                                   & 6M    & 0.36M & 94.3\%                                                 \\ 
\midrule
20                                                      & 14K   & 9.4\%                                                  & 3.5M  & 0.36M & 90\%                                                   \\ 
\midrule
100                                                     & 4K    & 2.7\%                                                  & 1M    & 0.36M & 73\%                                                   \\
\bottomrule
\end{tabular}
\caption{Illustration of the frequency-based vocabulary selection heuristic on a typical CNN-based document classification model (Section~\ref{sec:data_and_architecture}). \#Emb is the number of parameters in the word embedding matrix (256 dimensions), and \#CNN is that in the CNN model. }
\label{tab:embedding_ratio}
\end{table}
\section{Introduction}
\label{sec:intro}
Over the past decade, deep neural networks have become arguably the most popular model choice for a vast number of natural language processing (NLP) tasks and have constantly been delivering state-of-the-art results. Because neural network models assume continuous data, to apply a neural network on any text data, the first step is to vectorize the discrete text input with a word embedding matrix through look-up operation, which in turn assumes a pre-defined vocabulary set. For many NLP tasks, the vocabulary size can easily go up to the order of tens of thousands, which potentially makes the word embedding the largest portion of the trainable parameters. For example, a document classification task like AG-news~\cite{zhang2015character} can include up to 60K unique words, with the embedding matrix accounting for 97.6\% of the trainable parameters (\autoref{tab:embedding_ratio}), which leads to under-representation of the neural networks' own parameters.

Intuitively, using the full or very large vocabulary are neither economical, as it limits model applicability on computation- or memory-constrained scenarios~\cite{DBLP:conf/icml/YogatamaFDS15,DBLP:conf/acl/FaruquiTYDS15}, nor necessary, as many words may contribute little to the end task and could have been safely removed from the vocabulary. Therefore, how to select the best vocabulary is a problem of both theoretical and practical interests. Somewhat surprisingly, this \emph{vocabulary selection} problem is largely under-addressed in the literature: The \emph{de facto} standard practice is to do frequency-based cutoff~\cite{DBLP:conf/emnlp/LuongPM15,DBLP:conf/emnlp/Kim14}, and only retain the words more frequent than a certain threshold (\autoref{tab:embedding_ratio}). Although this simple heuristic has demonstrated strong empirical performance, its task-agnostic nature implies that likely it is not the optimal strategy for many tasks (or any task). Task-aware vocabulary selection strategies and a systematic comparison of different strategies are still lacking. 

In this work, we present the first systematic study of the vocabulary selection problem. Our study will be based on text classification tasks, a broad family of NLP tasks including document classification (DC), natural language inference (NLI), natural language understanding in dialog systems (NLU), etc. Specifically, we aim to answer the following questions:

\begin{enumerate}[topsep=0pt,itemsep=-1ex,partopsep=1ex,parsep=1ex]
    \item \emph{How important a role does the vocabulary selection algorithm play in text classification?}
    \item \emph{How to dramatically reduce the vocabulary size while retaining the accuracy?}
\end{enumerate}
The rest of the paper is organized as follows: We first formally define the vocabulary selection problem (\autoref{sec:problem_def}) and present a quantitative study on classification accuracy with different vocabulary selections to showcase its importance in the end task (\autoref{sec:monte_carlo}). We also propose two new metrics for evaluating the performance of vocabulary selection in text classification tasks (\autoref{sec:metrics}). We then propose a novel, task-aware vocabulary selection algorithm called \emph{Varitional Vocabulary Dropout (VVD)} (\autoref{sec:our_method}) which draws on the idea of variational dropout~\cite{kingma2015variational}: If we learn a dropout probability $p_w$ for each given word $w$ in the vocabulary $\mathbb{V}$ during the model training on a given task, the learned dropout probabilities $p_w$ will imply the importance of word $w$ to the end task and can, therefore, be leveraged for vocabulary selection. We propose to infer the latent dropout probabilities under a Bayesian inference framework. During test time, we select the sub vocabulary $\hat{\mathbb{V}}$ by only retaining words with dropout probability lower than a certain threshold. For any words deselected using VVD, we will simply regard them as a special token with null vector representation $[0, 0, \cdots, 0]$. Please note that our proposed algorithm needs to re-train a word embedding matrix, thus it is tangential to the research of pre-trained word embedding like Word2Vec~\cite{mikolov2013distributed} or Glove~\cite{pennington2014glove} though we can use them to initialize our embedding.

We conduct comprehensive experiments to evaluate the performance of VVD (\autoref{sec:experiments}) on different end classification tasks. Specifically, we compare against an array of strong baseline selection algorithms, including the frequency-based algorithm~\cite{DBLP:conf/emnlp/LuongPM15}, TF-IDF algorithm~\cite{ramos2003using}, and structure lasso algorithm~\cite{friedman2010note}, and demonstrate that it can consistently outperform these competing algorithms by a remarkable margin. To show that the conclusions are widely held, our evaluation is based on a wide range of text classification tasks and datasets with different neural networks including Convolutional Neural Network (CNN)~\cite{DBLP:conf/emnlp/Kim14}, Bi-directional Long-Short Term Memory (BiLSTM)~\cite{bahdanau2014neural} and Enhanced LSTM (ESIM)~\cite{DBLP:conf/acl/ChenZLWJI17}.
In summary, our contributions are three-fold:
\begin{enumerate}[topsep=0pt,itemsep=-1ex,partopsep=1ex,parsep=1ex]
    \item We formally define the vocabulary selection problem, demonstrate its importance, and propose new evaluation metrics for vocabulary selection in text classification tasks.
    \item We propose a novel vocabulary selection algorithm based on variational dropout by re-formulating text classification under the Bayesian inference framework. The code will be released in Github\footnote{\small \url{https://github.com/wenhuchen/Variational-Vocabulary-Selection.git}}.
    \item We conduct comprehensive experiments to demonstrate the superiority of the proposed vocabulary selection algorithm over a number of strong baselines.
\end{enumerate}

\begin{figure*}[ht!]
    \centering
    \includegraphics[width=0.95\linewidth]{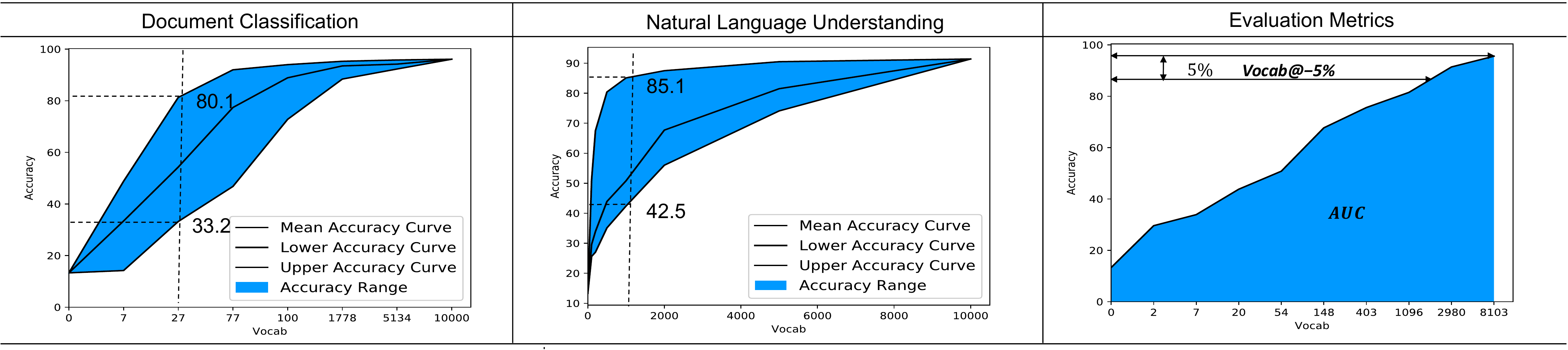}
    \caption{Monte-Carlo simulation on vocabulary selection. Left: CNN-based document classification on AG-news dataset. Middle: Natural language understanding on Snips dataset. Right: Metrics for vocabulary selection.}
    \label{fig:sampling}
\end{figure*}

\section{Vocabulary Selection}
\label{sec:vocabulary_selection}

\subsection{Problem Definition}
\label{sec:problem_def}

We now formally define the problem setting and introduce the notations for our problem. Conventionally, we assume the neural classification model vectorizes the discrete language input into a vector representation via an embedding matrix $W \in \mathbb{R}^{V*D}$, where $V$ denotes the size of the vocabulary, and $D$ denotes the vector dimension. The embedding is associated with a pre-defined word-to-index dictionary $\mathbb{V} = \{w_i:i| 1 \leq i \leq V\}$ where $w_i$ denotes a literal word corresponding to $i_{th}$ row in the embedding matrix. The embedding matrix $W$ covers the subset of a vocabulary of interests for a particular NLP task, note that the value of $V$ is known to be very large due to the rich variations in human languages. Here we showcase the embedding matrix size of a popular text classification model\footnote{\small \url{https://github.com/dennybritz/cnn-text-classification-tf}} on AG-news dataset~\cite{zhang2015character} in~\autoref{tab:embedding_ratio}. From which we can easily observe that the embedding matrix is commonly occupying most of the parameter capacity, which could be the bottleneck in many real-world applications with limited computation resources.

In order to alleviate such redundancy problem and make embedding matrix as efficient as possible, we are particularly interested in discovering the minimum row-sized embedding $\hat{W}$ to achieve nearly promising performance as using the full row-sized embedding $W$. More formally, we define the our problem as follows:
\begin{align}
\small
\label{eq:initial-1}
\begin{split}
\argmin_{\hat{W},\hat{\theta}} & \#Row(\hat{W})\\
s.t. \quad Acc(f_{\hat{\theta}}(x;\hat{W}), y) &-  Acc(f_{\theta}(x; W), y) \leq \epsilon
\end{split}
\end{align}
where \text{\#Row} is a the number of rows in the matrix $\hat{W}$, $f_{\theta}$ is the learned neural model with parameter $\theta$ to predict the class given the inputs $x$, $Acc$ is the function which measure accuracy between model prediction and $y$ (reference output), and $\epsilon$ is the tolerable performance drop after vocabulary selection. It is worth noting that here $\theta$ includes all the parameter set of the neural network except embedding matrix $W$. For each vocabulary selection algorithm $\mathcal{A}$, we propose to draw its characteristic curve $Acc(f_{\hat{\theta}}(x;\hat{W}), y) = g_{\mathcal{A}}(\#Row(\hat{W}))$ to understand the relationship between the vocabulary capacity and classification accuracy, which we call as (characteristic) accuracy-vocab curve throughout our paper.

\subsection{Importance of Vocabulary Selection}
\label{sec:monte_carlo}

In order to investigate the importance of the role played by the vocabulary selection algorithm, we design a Monte-Carlo simulation strategy to approximate accuracy's lower bound and upper bound of a given vocabulary size reached by a possible selection algorithm $\mathcal{A}$. More specifically, for a given vocabulary size of $\hat{V}$, there exist ${V \choose \hat{V}}$ algorithms which can select distinct vocabulary subset $\mathbb{\hat{V}}$ from the full vocabulary $\mathbb{V}$. Directly enumerating these possibilities are impossible, we instead propose to use a Monte-Carlo vocabulary selection strategy which can randomly pick vocabulary subset $\mathbb{\hat{V}}$ to simulate the possible selection algorithms by running it N times. After simulation, we obtain various point estimations $(Acc_1, \cdots, Acc_N|\hat{V})$ at each given $\hat{V}$ and depict the point estimates in~\autoref{fig:sampling} to approximately visualize the upper and lower bound of the accuracy-vocab curve. From~\autoref{fig:sampling}, we can easily observe that the accuracy range under a limited-vocabulary is extremely large, when the budget $\hat{V}$ increases, the gap gradually shrinks. For example, for document classification with a budget of 1000, a selection algorithm $\mathcal{A}$ can yield a potential accuracy ranging from $42.5$ to $85.1$, while for natural language understanding task with a budget of 27, a selection algorithm  $\mathcal{A}$ can yield a potential accuracy ranging from $33.2$ to $80.1$. Such a Monte-Carlo simulation study has demonstrated the significance of vocabulary selection strategy in NLP tasks and also implicate the enormous potential of an optimal vocabulary selection algorithm. 

\subsection{Evaluation Metrics}
\label{sec:metrics}

In order to evaluate how well a given selection algorithm $\mathcal{A}$ performs, we propose evaluation metrics as depicted in~\autoref{fig:sampling} by quantitatively studying its characteristic accuracy-vocab curve. These metrics namely Area Under Curve (AUC) and Vocab@-X\% separately measure the vocabulary selection performance globally and locally. Specifically, AUC computes enclosed area by the curve, which gives an overview of how well the vocabulary selection algorithm performs. In comparison, Vocab@-X\% computes the minimum vocabulary size required if X\% performance drop is allowed, which straightforwardly represents how large vocabulary is required to achieve a given accuracy. For the local evaluation metric, we mainly consider Vocab@-3\% and Vocab@-5\%. However, we observe that directly computing AUC lays too much emphasis on the large-vocabulary region, thus unable to represent an algorithm's selection capability under the low-vocabulary conditions. Therefore, we propose to take the logarithm of the vocabulary size and then compute the normalized enclosed area by:
\begin{align}
\small
    \begin{split}
        AUC = \frac{\int_{\hat{V}} Acc(\log(\hat{V}))d\log(\hat{V})}{\int_{\hat{V}} Acc(V)d\log(\hat{V})}
    \end{split}
\end{align}

It is worth noting that Vocab@-X\% takes value from range $[0, V]$ with smaller values indicate better performance. Since AUC is normalized by Acc(V), it takes value from range $[0, 1]$ regardless of the classification error.

\section{Our Method}
\label{sec:our_method}

\begin{figure}[htb]
    \centering
    \includegraphics[width=0.95\linewidth]{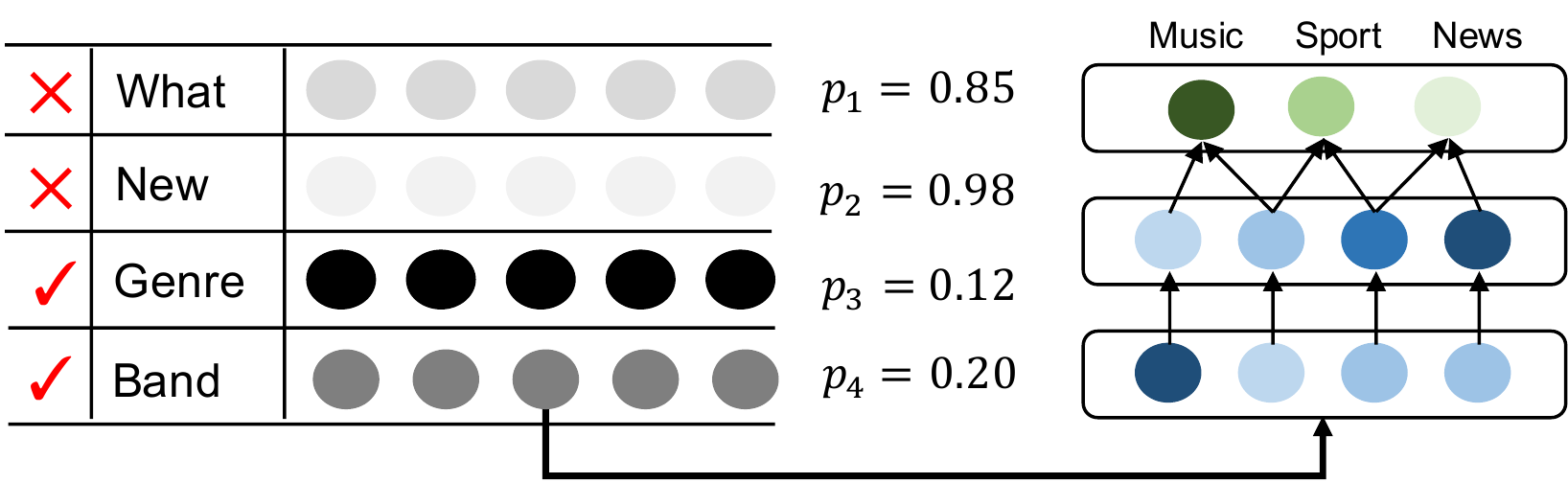}
    \caption{Variational dropout in classification models, ``New" and ``What" can be safely removed without harming performance due to large dropout probability.}
    \label{fig:variational}
\end{figure}

Inspired by DNN dropout~\cite{srivastava2014dropout,wang2013fast}, we propose to tackle the vocabulary selection problem from word-level dropout perspective, where we assume each word $w_i$ (an integer index) is associated with its characteristic dropout rate $p_i$, which represents the probability of being replaced with an empty placeholder, specifically, higher dropout probability indicates less loss suffered from removing it from the vocabulary. Hence, the original optimization problem in~\autoref{eq:initial-1} can be thought of as inferring the latent dropout probability vector $\mathbf{p} = [p_1, \cdots, p_V]$. The overview of our philosophy is depicted in~\autoref{fig:variational}, where we associate with each row of the embedding matrix a dropout probability and then re-train the complete system, which grasps how much contribution each word from the vocabulary makes to the end NLP task and remove those ``less contributory" words from the vocabulary without hurting the performance.

\subsection{Bernouli Dropout}
\label{sec:bernouli_dropout}
Here we first assume that the neural network vectorizes the discrete inputs with an embedding matrix $W$ to project given words $x$ into vector space $\mathbb{R}^D$, and then propose to add random dropout noise into the embedding input to simulate the dropout process as follows:
\begin{align}
\small
\begin{split}
    E(x|\mathbf{b}) = (\mathbf{b} \odot \onehot(x)) \cdot W 
\end{split}
\end{align}
where \textit{OneHot} is a function to transform a word $x$ into its one-hot form $\onehot(x) \in \mathbb{R}^V$, and $\mathbf{b} \in \mathbb{R}^V$ is the Bernouli dropout noise with $b_i \sim Bern(1-p_i)$. The embedding output vector $E(x|\mathbf{b})$ is computed with a given embedding matrix $W$ under a sampled Bernouli vector $\mathbf{b}$. In order to infer the latent Bernouli distribution with parameters $\mathbf{p}$ under the Bayesian framework where training pairs $(\mathbf{x}=x_1 \cdots x_n, y)$ are given as the evidence, we first define an objective function as $\Lagr(f_{\theta}(\mathbf{x}), y)$ and then derive its lower bound as follows (with $\bar{\mathbf{p}} = 1 - \mathbf{p}$):
\begin{align*}
\small
\begin{split}
    &\log \Lagr(f_{\theta}(\mathbf{x}), y) = \log \int_{\mathbf{b}} \Lagr(f_{\theta}(E(\mathbf{x}|\mathbf{b})), y) \mathcal{P}(\mathbf{b}) d\mathbf{b}\\
    \geq& \expect{\mathbf{b} \sim Bern(\mathbf{\bar{p}})} [\log \Lagr(f_{\theta}(E(\mathbf{x}|\mathbf{b})), y)] - KL(Bern(\mathbf{\bar{p}})||\mathcal{P}(\mathbf{b}))\\
    =& \Lagr(W; \theta)
\end{split}
\end{align*}
where $\mathcal{P}(\mathbf{b})$ is the prior distribution, and $Bern(\mathbf{\bar{p}})$ denotes the Bernouli approximate posterior with parameter $\mathbf{p}$. Here we use $E(\mathbf{x})$ as the simplied form of $\{ E(x_1), \cdots, E(x_n)\}$, we separate the text classification model's parameters $\theta$ with the embedding parameters $W$ and assume the classification model $f_{\theta}$ directly takes embedding $E$ as input.
\subsection{Gaussian Relaxation}
However, the Bernouli distribution is hard to reparameterize, where we need to enumerate $2^V$ different values to compute the expectation over the stochastic dropout vector $\mathbf{b}$. Therefore, we follow~\citet{wang2013fast} to use a continuous Gaussian approximation, where the Bernouli noise $\mathbf{b}$ is replaced by a Gaussian noise $\mathbf{z}$:
\begin{align}
\small
\begin{split}
    E(x|z) = (\mathbf{z} \odot \onehot(x)) \cdot W
\end{split}
\end{align}
where $\mathbf{z} \in \mathbb{R}^V$ follows Gaussian distribution $z_i \sim \mathcal{N}(1, \alpha_i=\frac{p_i}{1-p_i})$. It is worth noting that $\mathbf{\alpha}$ and $p$ are one-to-one corresponded, and $\alpha$ is a monotonously increasing function of $p$. For more details, please refer to~\citet{wang2013fast}. Based on such approximation, we can use $\alpha$ as dropout criteria, e.g. throw away words with $\alpha$ above a certain given threshold $\alpha_T$. We further follow~\citet{louizos2017bayesian,kingma2015variational,DBLP:conf/icml/MolchanovAV17} to re-interpret the input noise as the intrinsic stochasticity in the embedding weights $B$ itself as follows:
\begin{align}
\small
\begin{split}
    E(x|z) = \onehot(x) \cdot B
\end{split}
\end{align}
where $B \in \mathbb{R}^{V*D}$ follows a multi-variate Gaussian distribution $B_{ij} \sim \mathcal{N}(\mu_{ij}={W}_{ij}, \sigma_{ij}^2=\alpha_i W^2_{ij})$, where the random weights in each row has a tied variance/mean ratio $\alpha_i$. Thus, we re-write the evidence lower bound as follows:
\begin{align*}
\small
\begin{split}
    &\log \Lagr(f_{\theta}(\mathbf{x}), y)) = \log \int_{B} \Lagr(f_{\theta}(E(\mathbf{x}|z)), y)) \mathcal{P}(B) dB\\
    \geq& \expect{B \sim \mathcal{N}(\mu, \sigma)} [\log \Lagr(f_{\theta}(E(\mathbf{x}|z)), y)] - KL(\mathcal{N}(\mu, \sigma)||\mathcal{P}(B))\\
    =& \Lagr(B, \theta)
\end{split}
\end{align*}
where $\mathcal{P}(B)$ is the prior distribution and $\mathcal{N}(\mu, \sigma)$ denotes the Gaussian approximate posterior with parameters $\mu$ and $\sigma$. $\Lagr(B, \theta)$ is used as the relaxed evidence lower bound of marginal log likelihood $\log \Lagr(f_{\theta}(\mathbf{x}), y))$. Here, we follow~\citet{kingma2015variational,louizos2017bayesian} to choose the prior distribution $\mathcal{P}(B)$ as the ``improper log-scaled uniform distribution" to guarantee that the regularization term $D_{KL}(\mathcal{N}(\mu, \sigma)||\mathcal{P}(B))$ only depends on dropout ratio $\alpha$, i.e. irrelevant to $\mu$. Formally, we write the prior distribution as follows:
\begin{align}
\small
\begin{split}
    \mathcal{P}(\log |B_{ij}|) = const \rightarrow \mathcal{P}(|B_{ij}|) \propto \frac{1}{|B_{ij}|}
\end{split}
\end{align}
Since there exists no closed-form expression for such KL-divergence, we follow~\citet{louizos2017bayesian} to approximate it by the following formula with minimum variance:
\begin{align}
\small
\begin{split}
    D_{KL} &= -k_1 \sigma(k_2 + k_3\log \alpha) + \frac{1}{2} \log(1 + \frac{1}{\alpha}) + k_1\\
    k_1 &=0.63576 \quad k_2=1.87320 \quad k_3=1.48695
\end{split}
\end{align}
By adopting the improper log-uniform prior, more weights are compressed towards zero, and the KL-divergence is negatively correlated with dropout ratio $\alpha$. Intuitively, the dropout ratio $\alpha_i$ is an redundancy indicator for $i_{th}$ word in the vocabulary, with larger $\alpha_i$ meaning less performance loss caused by dropping $i_{th}$ word. During training, we use re-parameterization trick~\cite{kingma2013auto} to sample embedding weights from the normal distribution to reduce the Monte-Carlo variance in Bayesian training.
\subsection{Vocabulary Selection}
After optimization, we can obtain the dropout ratio $\alpha_i$ associated with each word $w_i$. We propose to select vocabulary subset based on the dropout ratio by using a threshold $\alpha_T$. Therefore, the remaining vocabulary subset is described as follows:
\begin{align}
\small
\begin{split}
        \mathbb{\hat{V}} = \{w_i \in V | \alpha_i < \alpha_T\}
\end{split}
\end{align}
where we use $\mathbb{\hat{V}}$ to denote the subset vocabulary of interest, by adjusting $\alpha_T$ we are able to control the selected vocabulary size.

\section{Experiments}
\label{sec:experiments}
We compare the proposed vocabulary selection algorithm against several strong baselines on a wide range of text classification tasks and datasets.

\begin{table*}
\small
\centering
\begin{tabular}{llllll} 
\toprule
Datasets & Task                                                                                               & Description                              & \#Class & \#Train & \#Test  \\
\midrule
ATIS-flight~\cite{tur2010left} & \multirow{2}{*}{\begin{tabular}[c]{@{}l@{}}NLU\end{tabular}}  & Classify Airline Travel dialog           & 21      & 4,478   & 893     \\
Snips~\cite{DBLP:journals/corr/abs-1805-10190}       &                         & Classify inputs to personal voice assistant & 7       & 13,084  & 700     \\
\midrule
AG-news~\cite{zhang2015character}     &  \multirow{4}{*}{\begin{tabular}[c]{@{}l@{}}DC\end{tabular}}  & Categories: World, Sports, etc  & 4 & 120,000 & 7,600   \\
DBPedia~\cite{lehmann2015dbpedia}     &                      & Categories: Company, Athlete, Album, etc & 14      & 560,000 & 70,000  \\
Sogou-news~\cite{zhang2015character}  &                         & Categories: Sports, Technology, etc      & 5       & 450,000 & 60,000  \\
Yelp-review~\cite{zhang2015character} &                         & Categories: Review Ratings (1-5)         & 5       & 650,000 & 50,000  \\ 
\midrule
SNLI~\cite{DBLP:conf/emnlp/BowmanAPM15}        &  \multirow{2}{*}{\begin{tabular}[c]{@{}l@{}}NLI\end{tabular}}                   & Entailment: Contradict, Neutral,Entail   & 3       & 550,152 & 10,000  \\
MNLI~\cite{DBLP:conf/naacl/WilliamsNB18}        & & Multi-Genre Entailment   & 3       & 392,702 & 10,000  \\
\bottomrule
\end{tabular}
\caption{An overview of different datasets under different classification tasks including description and sizes.}
\label{tab:dataset}
\end{table*}
\subsection{Datasets \& Architectures}
\label{sec:data_and_architecture}
The main datasets we are using are listed in~\autoref{tab:dataset}, which provides an overview of its description and capacities. Specifically, we follow~\cite{zhang2015character,goo2018slot,DBLP:conf/naacl/WilliamsNB18} to pre-process the document classification datasets, natural language understanding dataset and natural language inference dataset. We exactly replicate their experiment settings to make our method comparable with theirs. Our models is implemented with TensorFlow~\cite{tensorflow2015-whitepaper}.
\begin{figure*}[thb]
    \centering
    \includegraphics[width=0.90\linewidth]{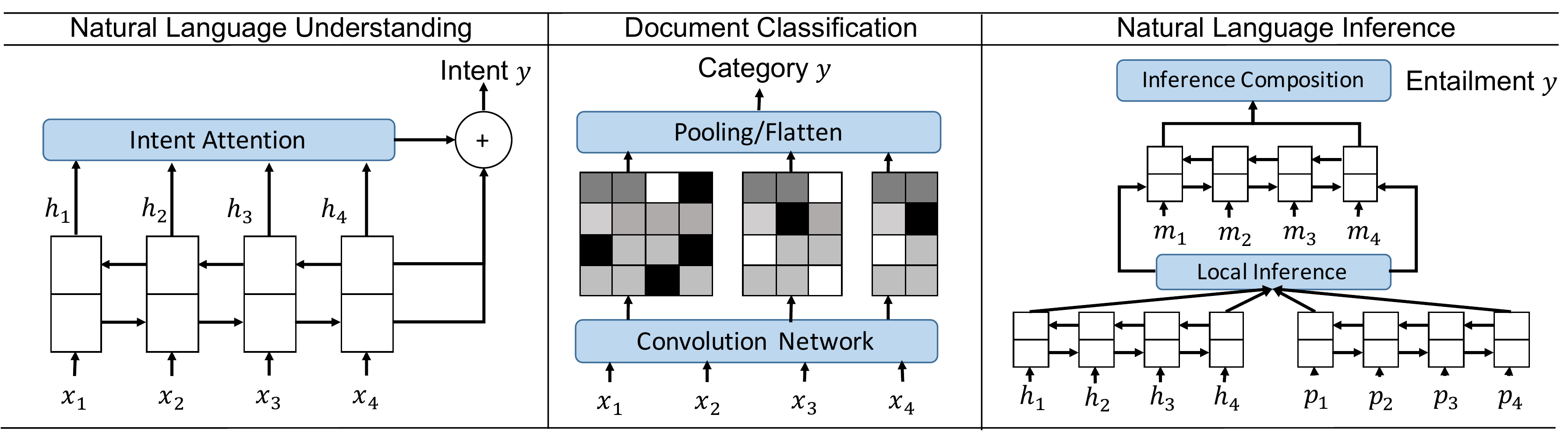}
    \caption{The neural network architecture overview of different NLP tasks. }
    \label{fig:architect}
\end{figure*}
In order to evaluate the generalization ability of VVD selection algorithm in deep learning architectures, we study its performance under different established architectures (depicted in~\autoref{fig:architect}). In natural language understanding, we use the most recent attention-based model for intention tracking~\cite{goo2018slot}, this model first uses BiLSTM recurrent network to leverage left-to-right and right-to-left context information to form the hidden representation, then computes self-attention weights to aggregate the hidden representation and predicts user intention. In document classification, we mainly follow the CNN architecture~\cite{DBLP:conf/emnlp/Kim14} to extract n-gram features and then aggregate these features to predict document category. In natural language inference, we follow the popular ESIM architecture~\citep{DBLP:conf/naacl/WilliamsNB18, DBLP:conf/acl/ChenZLWJI17} using the Github implementation\footnote{\small\url{https://github.com/coetaur0/ESIM}}. In this structure, three main components input encoding, local inference modeling, and inference composition are used to perform sequential inference and composition to simulate the interaction between premises and hypothesis. Note that, we do not apply the syntax-tree based LSTM proposed in~\citep{DBLP:conf/acl/ChenZLWJI17} because we lost the parse tree~\cite{klein2003fast} after the vocabulary compression, instead, we follow the simpler sequential LSTM framework without any syntax parse as input. Besides, the accuracy curve is obtained using the publicly available test split rather than the official online evaluation because we need to evaluate lots of times at different vocabulary capacity.

\begin{table*}[thb]
\centering
\small
\begin{tabular}{lcclccc} 
\toprule
Datasets / Reported Accuracy         & Accuracy & Vocab                & Methods             & AUC         & Vocab@-3\% & Vocab@-5\%  \\ 
\midrule
\multirow{4}{*}{Snips / 96.7~\cite{DBLP:conf/interspeech/LiuL16}} & 95.9       &  \multirow{4}{*}{11000} & Frequency       &  77.4    & 81         & 61          \\
                       & 95.9       &                        & TF-IDF           &   77.6   & 81           & 62          \\
                       & 95.6       &                        & Group Lasso       &   82.1   & 77           & 52\\
                       & 96.0 &                        & VVD & \textbf{82.5}  & \textbf{52}  & \textbf{36}         \\ 
\midrule
\multirow{4}{*}{ATIS-Flight / 94.1~\cite{goo2018slot}}  & 93.8       &  \multirow{4}{*}{724}    & Frequency      &  70.1   & 33           & 28    \\
                       & 93.8       &                        & TF-IDF           &  70.5  & 34          & 28\\
                       & 93.8       &                        & Group Lasso      &  72.9  & 30           & 26\\
                       & 94.0       &                        & VVD & \textbf{74.8}  & \textbf{29}  & \textbf{26}  \\
\midrule
\midrule
\multirow{5}{*}{AG-news / 91.1~\cite{zhang2015character}}     & 91.6       &     \multirow{4}{*}{61673} & Frequency    &  67.1 &    2290        & 1379\\
                             & 91.6       &                        & TF-IDF          &     67.8    &    2214     &    1303 \\

                             & 91.2       &                        & Group Lasso         &  68.3   &    1867     &  1032 \\
                             & 91.6       &                        & VVD & \textbf{70.5}  &  \textbf{1000}  & \textbf{673}   \\ 
\midrule
\multirow{4}{*}{DBPedia / 98.3~\cite{zhang2015character}}     & 98.4       &  \multirow{4}{*}{563355} & Frequency          &  69.7 &    1000    & 743            \\
                             & 98.4       &                        & TF-IDF           &   71.7    & 1703           & 804      \\

                             & 97.9       &                        & Group Lasso         &     71.9   &     768        &    678        \\
                             & 98.5       &               & VVD & \textbf{72.2}  & \textbf{427}   & \textbf{297}   \\ 
\midrule
\multirow{4}{*}{Sogou-news / 95.0~\cite{zhang2015character}}  & 93.7       & \multirow{4}{*}{254495}  & Frequency     & 70.9  & 789 & 643 \\
                             & 93.7       &                        & TF-IDF           & 71.3    & 976  &  776      \\
                             & 93.6       &                        & Group Lasso      & 73.4  &  765 &  456 \\
                             & 94.0       &                        & VVD & \textbf{75.5}  & \textbf{312}   & \textbf{196}   \\ 
\midrule
\multirow{4}{*}{Yelp-review / 58.0~\cite{zhang2015character}} & 56.3       &    \multirow{4}{*}{252712} & Frequency           &  74.0        &      1315      &     683        \\
                             & 56.3       &                        & TF-IDF           &     74.1      &     1630       &    754       \\

                             & 56.5       &                        & Group Lasso         &    75.4       &  934       &     463     \\
                             & \textbf{57.4}       &                        & VVD & \textbf{77.9}      & \textbf{487} & \textbf{287}   \\
\midrule
\midrule
\multirow{4}{*}{SNLI / 86.7~\cite{DBLP:conf/naacl/WilliamsNB18}} & 84.1     &  \multirow{4}{*}{42392} & Frequency           &   72.2   &   2139      &  1362 \\
                      & 84.1     &                        & TF-IDF              &  72.8    &    2132   &      1429 \\
                      & 84.6     &                        & Group Lasso         &  73.6    &    1712  &      1093 \\
                      & \textbf{85.5}     &                        & VVD & \textbf{75.0} & \textbf{1414} & \textbf{854}  \\
\midrule
\multirow{4}{*}{MNLI / 72.3~\cite{DBLP:conf/naacl/WilliamsNB18}} & 69.2     &  \multirow{4}{*}{100158} & Frequency           &   78.5   & 1758         &  952        \\
& 69.2     &                        & TF-IDF              &   78.7  &   1656      & 934 \\
& 70.1     &                        & Group Lasso         &   79.2    & 1466         &  711 \\
& \textbf{71.2}     &                        & VVD &  \textbf{80.1} &  \textbf{1323}  & \textbf{641} \\
\bottomrule
\end{tabular}
\caption{Experimental Results on various NLP tasks and datasets on the proposed metrics in~\autoref{sec:metrics}. Bold accuracy means the result is statistically significantly better than the competitors. }
\label{tab:result}
\end{table*}
\subsection{Baselines}
Here we mainly consider the following baselines:
\paragraph{Frequency-based (task-agnostic)}
This approach is already extensively talked about in~\autoref{sec:intro}, its basic idea is to rank the word based on its frequency and then set a threshold to cut off the long tail distribution.
\paragraph{TF-IDF (task-agnostic)} This algorithm views the vocabulary selection as a retrieval problem~\cite{ramos2003using}, where term frequency is viewed as the word frequency and document frequency is viewed as the number of sentences where such word appears. Here we follow the canonical TF-IDF approach to compute the retrieval score as follows:
\begin{align}
\small
\begin{split}
    tfidf(w, D) = tf(w)^\lambda * (\log \frac{N}{n_w})^{1 - \lambda}
\end{split}
\end{align}
where $tf(w)$ denotes the word frequency, $\lambda$ is the balancing factor, $N$ denotes the number of sentences and $n_w$ denotes the number of sentences in which $w$ appears. We rank the whole vocabulary based on the $tfidf$ and cut off at given threshold.  
\paragraph{Group Lasso (task-aware)} This baseline aims to find intrinsic sparse structures~\cite{liu2015sparse,park2016faster,wen2016learning} by grouping each row of word embedding. The regularization objective is described as follows, which aims at finding the row-wise sparse structure:
\begin{align}
\small
\begin{split}
    \mathcal{L}_{reg} = \sum_i(\sum_j W_{ij}^2)^{\frac{1}{2}}
\end{split}
\end{align}
After optimized with the above regularization, we use a threshold-based selection strategy on the row-norm of embedding matrix, the selected vocabulary is described as $\mathbb{\hat{V}} = \{w_i \in \mathbb{V}| ||W_{i}||_2 > \beta_T \}$, where $\beta_T$ is the threshold.
\begin{figure*}[thb]
    \centering
    \includegraphics[width=0.95\linewidth]{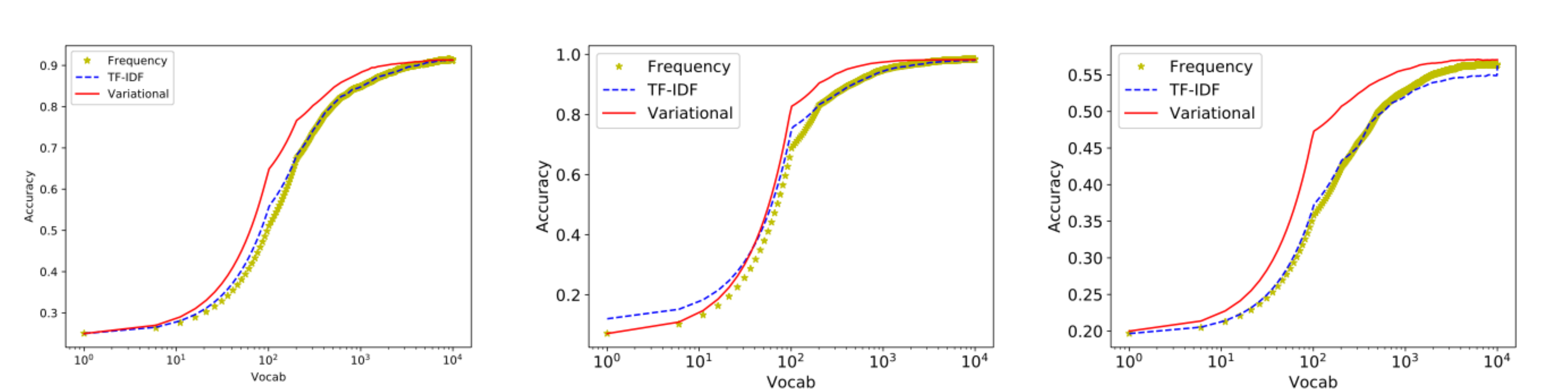}
    \caption{The accuracy-vocab curve of VVD, TF-IDF and frequency-based baseline, the datasets used are AG-news, DBPedia and Yelp-review respectively.}
    \label{fig:curve}
\end{figure*}
\begin{figure}[thb]
    \centering
    \includegraphics[width=1.0\linewidth]{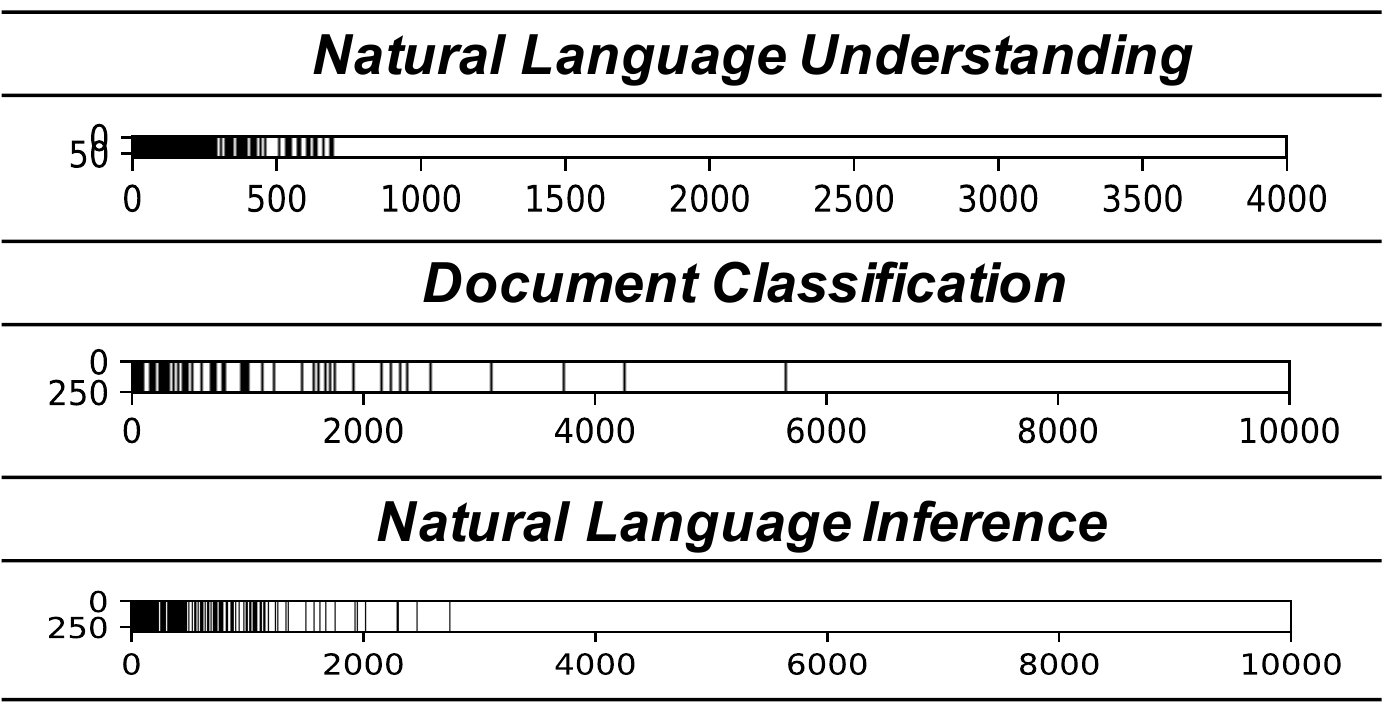}
    \caption{The vocabulary selection spectrum of our proposed VVD algorithm on different NLP tasks. }
    \label{fig:embedding}
\end{figure}
\begin{figure}[thb]
    \centering
    \includegraphics[width=1.0\linewidth]{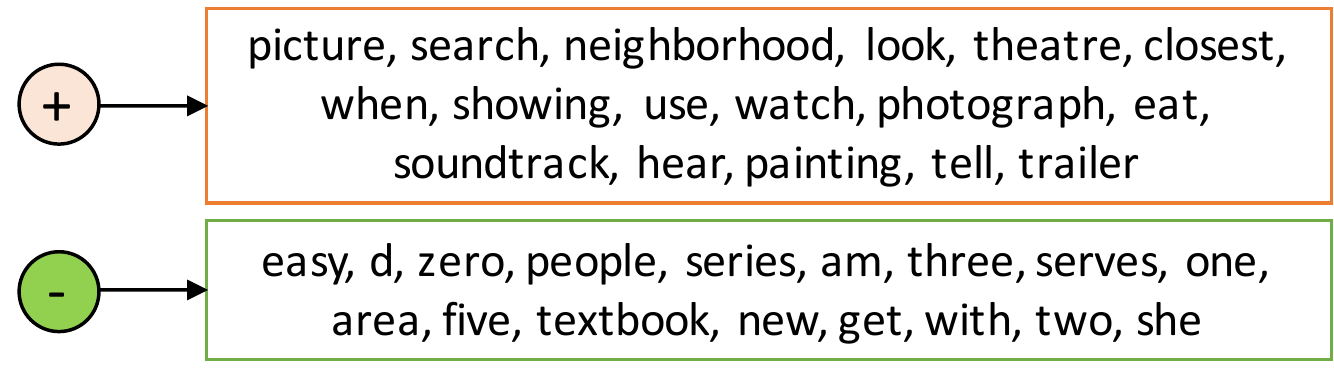}
    \caption{The vocabulary selected (+)/unselected (-) by VVD compared to frequency-based baseline.}
    \label{fig:wordcloud}
\end{figure}
\begin{figure}[thb]
    \centering
    \includegraphics[width=1.0\linewidth]{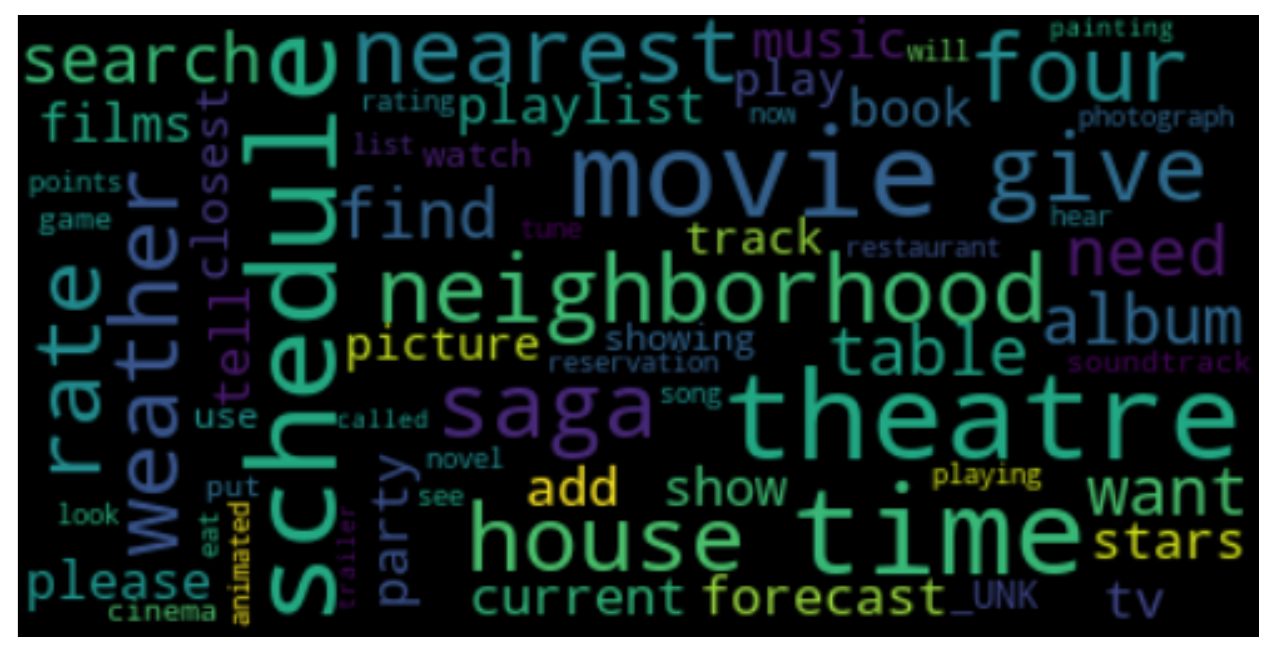}
    \caption{The vocabulary cloud of Snips NLU dataset.}
    \label{fig:wordcloud_real}
\end{figure}
\subsection{Experimental Results}
Here we demonstrate our results in natural language understanding, document classification, and natural language inference separately in~\autoref{tab:result}. From these tables, first of all, we can observe that VVD is able to maintain or even improve the reported accuracy on DC and NLU tasks, the accuracy of VVD is reported under dropping out the words with dropout rate larger than $0.95$. The exception is in NLI~\cite{DBLP:conf/naacl/WilliamsNB18}, where the common approach uses GloVe~\cite{pennington2014glove} for initialization, and we use random initialization, which makes our model fall slightly behind. It is worth noting that Frequency-based/TF-IDF methods are based on the model trained with cross entropy, while both Group-Lasso and VVD modify the objective function by adding additional regularization. It can be seen that VVD is performing very similar to the baseline models on DC and NLU tasks, while consistently outperforming the baseline methods (with random initialized embedding) on more challenging NLI and Yelp-Review tasks, that said, VVD can also be viewed as a generally effective regularization technique to sparsify features and alleviate the over-fitting problem in NLP tasks. In terms of the vocabulary selection capability, our proposed VVD is demonstrated to outperform the competing algorithms in terms of both AUC and Vocab@-X\% metrics consistently over different datasets as shown in~\autoref{tab:result}. In order to better understand the margin between VVD and frequency-based method, we plot their accuracy-vocab curves in~\autoref{fig:curve}, from which we can observe that the accuracy curves start from nearly the same accuracy with the full vocabulary, by gradually decreasing the budget $\hat{V}$, VVD decreases at a much lower rate than the competing algorithms, which clearly reflects its superiority under limited-budget scenario. From the empirical result, we can conclude that: 1) the retrieval-based selection algorithm can yield marginal improvement over the AUC metric, but the vocab@-X\% metric deteriorates.  2) group-lasso and VVD algorithm directly considers the connection between each word and end classification accuracy; such task-awareness can greatly in improving both evaluation metrics. Here we show that NLU datasets are relatively simpler, which only involves detecting key words from human voice inputs to make decent decisions, a keyword vocabulary within 100 is already enough for promising accuracy. For DC datasets, which involve better inner-sentence and inter-sentence understanding, hundred-level vocabulary is required for most cases. NLI datasets involve more complicated reasoning and interaction, which requires a thousand-level vocabulary. 
\paragraph{Case Study} To provide an overview of what words are selected, we depict the selection spectrum over different NLP tasks in~\autoref{fig:embedding}, from which we observe that most of the selected vocabulary are still from the high-frequency area to ensure coverage, which also explains why the frequency-based algorithm is already very strong. Furthermore, we use the Snips dataset~\cite{DBLP:journals/corr/abs-1805-10190} to showcase the difference between the vocabularies selected by VVD and by frequency-based baseline. The main goal of this dataset is to understand the speaker's intention such as ``BookRestaurant", ``PlayMusic", and ``SearchLocalEvent". We show the selected/unselected words by our algorithm in~\autoref{fig:wordcloud} under a vocabulary budget of 100, it is observed that many non-informative but frequent functional words like ``get", ``with", and ``five" are unselected while more task-related but less frequent words like ``neighborhood", ``search", ``theatre'' are selected. More vividly, we demonstrate the word cloud of the selected vocabulary of Snips~\cite{DBLP:journals/corr/abs-1805-10190} in~\autoref{fig:wordcloud_real}.

\subsection{Discussion}
Here we will talk about some potential issues posed when training and evaluating VVD.
\paragraph{Training Speed}
Due to the stochasticity of VVD, the training of text classification takes longer than canonical cross entropy objective. More importantly, we observe that with the increase the full vocabulary size, the convergence time of VVD also increases sub-linearly but the convergence time of Cross Entropy remains quite consistent. We conjecture that this is due to the fact that the VVD algorithm has a heavier burden to infer the drop out the probability of the long tail words. Therefore, we propose to use a two-step vocabulary reduction to dramatically decrease VVD's training time, in the first step, we cut off the rare words without having any harm on the final accuracy, then we continue training with VVD on the shrunk vocabulary. By applying such a hybrid methodology, we are able to decrease the training time dramatically.
\paragraph{Evaluation Speed}
As we know, at each vocabulary point, the network needs to perform once evaluation on the whole test set. Therefore, it is not practical to draw each vocabulary size from 1 to V and perform V times of evaluation. Given the limited computational resources, we need to sample some vocabulary size and estimate the area under curve relying on only these points. Uniformly sampling the data points are proved wasteful, since when the accuracy curve will converge to a point very early, most of the sampled point is actually getting equivalent accuracy. Therefore, we propose to increase the interval exponentially to cover more samples at extremely low vocabulary size. For example, given the total vocabulary of 60000, the interval will be split into 1, 2, 4, 8, 24, 56, ..., 60K. Using such sampling method achieve a reasonably accurate estimation of ROC with only $O(log(|V|))$ sample points, which is affordable under many cases. 

\section{Related Work}
\paragraph{Neural Network Compression}
In order to better apply the deep neural networks under limited-resource scenarios, much recent research has been performed to compress the model size and decrease the computation resources. In summary, there are mainly three directions, weight matrices approximation~\cite{DBLP:journals/corr/LeJH15,DBLP:conf/ijcnn/TjandraSN17}, reducing the precision of the weights~\cite{hubara2017quantized,han2015deep} and sparsification of the weight matrix~\cite{wen2016learning}. Another group of sparsification relies on the Bayesian inference framework~\cite{DBLP:conf/icml/MolchanovAV17,neklyudov2017structured,louizos2017bayesian}. The main advantage of the Bayesian sparsification techniques is that they have a small number of hyperparameters compared to pruning-based methods. As stated in~\cite{chirkova2018bayesian}, Bayesian compression also leads to a higher sparsity level~\cite{DBLP:conf/icml/MolchanovAV17,neklyudov2017structured,louizos2017bayesian}. Our proposed VVD is inspired by these predecessors to specifically tackle the vocabulary redundancy problem in NLP tasks. 
\paragraph{Vocabulary Reduction}
An orthogonal line of research for dealing similar vocabulary redundancy problem is the character-based approaches to reduce vocabulary sise~\cite{kim2016character,zhang2015character,DBLP:conf/acl/Costa-JussaF16,DBLP:journals/tacl/LeeCH17}, which decomposes the words into its characters forms for better handling open world inputs. However, these approaches are not applicable to character-free languages like Chinese and Japanese. Moreover, splitting words into characters incurs potential lose of word-level surface form, and thus needs more parameters at the neural network level to recover it to maintain the end task performance~\cite{zhang2015character}, which contradicts with our initial motivation of compressing the neural network models for computation- or memory-constrained scenarios.  

\section{Conclusion}
In this paper, we propose a vocabulary selection algorithm which can find sparsity in the vocabulary and dynamically decrease its size to contain only the useful words. Through our experiments, we have empirically demonstrated that the commonly adopted frequency-based vocabulary selection is already a very strong mechanism, further applying our proposed VVD can further improve the compression ratio. However, due to the time and memory complexity issues, our algorithm and evaluation are more suitable for classification-based application. In the future, we plan to investigate broader applications like summarizaion, translation, question answering, etc.

\section{Acknowledgement}
The authors would like to thank the anonymous reviewers for their thoughtful comments. This research was sponsored in part by NSF 1528175. Any opinions, findings, and conclusions or recommendations expressed in this material are those of the author(s) and do not necessarily reflect the views of the National Science Foundation.

\clearpage
\bibliography{naaclhlt2019}
\bibliographystyle{acl_natbib}

\end{document}